\documentclass[letterpaper]{article}
\usepackage[final]{neurips_2019}
\usepackage{enumitem}
\usepackage{mathtools}
\usepackage{multirow}
\usepackage{microtype}
\usepackage{wrapfig}
\usepackage{subfigure}
\usepackage{adjustbox}

\title{The Case for Evaluating Causal Models \\ Using Interventional Measures and Empirical Data}
\author{Amanda Gentzel, Dan Garant, and David Jensen\\
College of Information and Computer Sciences\\
University of Massachusetts Amherst}

\begin{document}
\setlength{\belowcaptionskip}{-12pt}

\maketitle

\begin{abstract}
Causal inference is central to many areas of artificial intelligence, including complex reasoning, planning, knowledge-base construction, robotics, explanation, and fairness. An active community of researchers develops and enhances algorithms that learn causal models from data, and this work has produced a series of impressive technical advances.  However, \emph{evaluation techniques} for causal modeling algorithms have remained somewhat primitive, limiting what we can learn from experimental studies of algorithm performance, constraining the types of algorithms and model representations that researchers consider, and creating a gap between theory and practice.  We argue for more frequent use of evaluation techniques that examine \emph{interventional measures} rather than structural or observational measures, and that evaluate those measures on \emph{empirical data} rather than synthetic data.  We survey the current practice in evaluation and show that the techniques we recommend are rarely used in practice. We show that such techniques are feasible and that data sets are available to conduct such evaluations.  We also show that these techniques produce substantially different results than using structural measures and synthetic data.
\end{abstract}

\section{Introduction}
Evaluation is central to research in artificial intelligence and machine learning \citep{cohen1995empirical,Langley2011changing}.  How we evaluate algorithms determines our perception of the relative effectiveness and usefulness of different approaches, and this knowledge guides choices about future research directions.  As \citet{Cohen1989} explained three decades ago: ``Ideally, evaluation should be a mechanism by which AI progresses both within and across individual research projects. It should be something we do as individuals to help our own research and, more importantly, on behalf of the field.''

As fields develop, protocols for evaluation need to develop alongside them.  In this paper, we offer an empirical analysis of the set of techniques typically used to evaluate algorithms for learning causal models, and we show that this set could be substantially enhanced.  The ultimate goal of most algorithms for causal inference is to learn models capable of accurately estimating the effects of interventions in real-world systems.  With this goal in mind, we would like to evaluate algorithms by comparing their estimates to actual interventional effects on data produced by a real-world system.  In practice, though, many evaluations fall short of this ideal, most frequently using only synthetic data and structural or observational measures.  Without the use of \textit{empirical data}, our evaluations produce little information about whether our algorithms generalize to real-world systems, and this greatly reduces their likelihood of widespread adoption by others outside of the field.  Without the use of \textit{interventional measures}, our evaluations produce little information about whether learned models will accurately estimate the effects of interventions, limiting their real-world utility.  

Note that we do not argue for \emph{replacing} the prevailing techniques for evaluation.  These techniques have substantial value, both in assessing overall performance and in allowing fine-grained experiments to diagnose specific performance issues.  Rather, we argue for \emph{augmenting} the current suite of evaluation techniques to gather experimental evidence that the prevailing techniques cannot.
We also do not contend that interventional measures and empirical data are entirely absent from current studies.  A very small minority of recent studies use these techniques in combination.  Rather, we argue that interventional measures and empirical data should be used routinely, and should be used in combination, for any serious study of algorithms for learning causal models. Indeed, the conclusions of most studies that lack such evaluation techniques should be considered exploratory and would benefit from additional evaluation.

We make the following contributions:

\begin{enumerate}[label=\textbf{C\arabic*}, leftmargin=*]
 
\item \textbf{Decomposition of Evaluation Techniques.} We decompose evaluation techniques into three interacting components: the data source, the algorithm, and the evaluation measure, allowing for a modular discussion of the interacting components of an evaluation.

\item \textbf{Survey of Current Techniques.} We provide a detailed survey of recent literature in causal modeling to provide a quantitative understanding of current evaluation practices.

\item \textbf{Critique of Current Practice.}  We provide evidence that increased adoption of both empirical data and interventional measures would be beneficial to the community.

\end{enumerate}

\section{Survey of Current Techniques}
\label{sec:survey}
To assess how frequently different evaluation techniques are used in practice, we surveyed recent computer science publications on causal inference.  We collected papers from the past five UAI, NeurIPS, AAAI, ICML, and KDD conferences, as well as causality workshops held at UAI.  We examined papers whose titles contained the terms `cause', `causal', or `causality' and then narrowed this selection of papers to those that describe, propose, or evaluate a causal modeling algorithm. This resulted in a final set of 111 papers, of which 82\% (91) reported any sort of evaluation.\footnote{When reporting survey results, we follow each percentage with a parenthesized number representing the raw count.  The denominator for percentages is 91, except where otherwise noted.}  Citations to all 111 papers are provided in the Supplementary Material.

The counts of papers included in the final survey are shown in Table \ref{tbl:survey_counts}.  While some relevant papers may fall outside of our search parameters, this approach captures a reasonably representative sample of recent work within computer science on causal modeling, allowing us to infer which techniques are used in practice and how frequently these techniques are used.


\begin{table}[h]
\begin{adjustbox}{width=1\columnwidth, center}
\small
\begin{minipage}[b]{.5\linewidth}
\caption{Papers included in the survey}
\label{tbl:survey_counts}
\renewcommand{\arraystretch}{1.18}
\addtolength{\tabcolsep}{-2pt}
\begin{tabular}{r llllll c}
\textbf{Venue} & 2014 & 2015 & 2016 & 2017 & 2018 & \textbf{Total}\\
\hline
UAI & 2 & 3 & 5 & 3 & 7 & 20\\
  NeurIPS & 3 & 5 & 4 & 6 & 13 & 31\\
  AAAI & 1 & 6 & 2 & 4 & 5 & 18\\
  ICML & 1 & 5 & 1 & 3 & 5 & 15\\
  KDD & 0 & 2 & 3 & 0 & 2 & 7\\
  UAI-W & 2 & 2 & 4 & 3 & 9 & 20\\
  \hline
  \textbf{Total} & 9 & 23 & 19 & 19 & 41 & 111\\
\end{tabular}
\end{minipage}
\hspace{.5cm}
\begin{minipage}[b]{.45\linewidth}
\caption{Number of papers using different evaluation measures}
\label{tbl:survey_results}
\renewcommand{\arraystretch}{1.13}
\addtolength{\tabcolsep}{-2pt}
\begin{tabular}{cr|cc}
\multicolumn{2}{c}{} & \multicolumn{2}{c}{\textbf{Data Sources}}\\
\multicolumn{2}{c}{} & Synthetic & Empirical\\
\cline{3-4}
\multirow{4}{*}{{\rotatebox{90}{\parbox{0.3\columnwidth}{\raggedleft \textbf{Evaluation Measures}}}}}&
 Structural        & 44 & 23 \\
&Observational     & 22 & 14 \\
&Interventional    & 11 & 6  \\
&Visual Inspection & 0  & 19 \\
\end{tabular}
\end{minipage}
\end{adjustbox}
\end{table}

\subsection{Survey Results}

For ease of exposition, we decompose evaluation techniques into three components: (1) the data source; (2) the algorithm under evaluation; and (3) the evaluation measure.  These dimensions are highly dependent---a choice of one can determine feasible choices for the others.  For example, models learned from observational macro-economic data often cannot be compared against a known structure because there exists no ground truth, and models consisting only of non-parameterized structure cannot be compared to interventional effects because the models cannot produce such estimates.

\textbf{Data Sources.} The surveyed papers used a wide range of data sources, but they fall into two broad categories: synthetic and empirical.  We categorized data as empirical when it was collected from a ``real world'' system, whether that was a randomized clinical trial, a global financial system, or user interaction with a website.  The important distinction is that empirical data was collected from a process or a system that exists for some purpose beyond scientific research.
Synthetic data includes anything else, including data generated from a randomly instantiated directed graphical model or from a simulation intended to reflect a real-world system.  In our survey, we found many examples of both, and while synthetic data is used more frequently, both are still common.  81\% (74) of papers surveyed used synthetic data, 67\% (61) used empirical data, and 48\% (44) used both.


\textbf{Algorithms.} The algorithm under evaluation is not part of the evaluation technique \emph{per se}, but aspects of the algorithm strongly influence how evaluation can, and should, be performed.  Algorithms fall into two broad categories, bivariate and multivariate, based on the number of variables they consider, although there are many variants.  

Some bivariate algorithms infer only the direction of effect (whether $A$ causes $B$ or $B$ causes $A$).  Others estimate the magnitude of effect between treatment and outcome, while adjusting for the effects of a number of covariates.  Bivariate methods include Granger causality analysis \citep{granger1969investigating}, additive noise models \citep{peters2014causal}, and analyses that use the potential outcomes framework \citep{rubin2005causal}. The most common variety of multivariate algorithm learns a directed acyclic graph (DAG).  Multivariate algorithms are significantly more prevalent in the data, accounting for 60\% (55/111) of papers surveyed.  Bivariate algorithms account for 30\% (34/111) of papers surveyed, split between those focused on orientation (10\%), magnitude of effect (15\%), or both (5\%).  The remaining papers in the survey fall in between, including those that aim to determine the joint effect of multiple treatment variables on a single outcome.

\textbf{Evaluation Measures.} At the heart of any evaluation technique is a measure of performance.  At a high level, evaluation measures fall into two categories: structural and distributional.  Structural measures include all measures designed to assess whether the structure (including both existence of edges and edge orientation) learned by the algorithm matches the ground truth.  Structural measures include structural Hamming distance (SHD), precision, recall, F1-score, true-positive rate, area under the ROC curve (AUROC), and structural intervention distance (SID)~\citep{peters2015structural}.

Distributional measures capture how well the algorithm can estimate quantitative dependence.  Such measures can be further subdivided into observational and interventional measures.  Observational measures compare the learned distribution with an observational ground truth (i.e. probability queries which do not involve a \emph{do} operator).  This could be a measure of individual edge strengths in a directed graphical model or a measure of the error when predicting a given outcome variable.  Interventional measures, on the other hand, compare the learned distribution to ground truth obtained through intervention.  Common interventional measures include  KL-divergence, total variation distance, and measures of average and conditional treatment effect.


Of the types of evaluation measures, structural measures are the most common, being used in 55\% (50) of papers surveyed. Distributional measures are slightly less common, being used in 46\% (42) of papers.  The vast majority of the distributional measures used, however, are observational rather than interventional; observational measures are used in 32\% (29) of papers, while interventional measures are used in only 14\% (13).

The choice of evaluation measure depends on both the data generating process and type of algorithm, which is reflected in our survey.  When synthetic data is evaluated, structural measures are used 59\% (44/74) of the time. However, when empirical data is evaluated, structural measures are used only 38\% (23/61) of the time, since empirical data is less likely to have ground truth.  This lack of ground truth sometimes prevents any significant evaluation for techniques using empirical data---26\% (16/61) of empirical evaluations used only visual inspection of the results, with no ground truth.  Table \ref{tbl:survey_results} summarizes the interaction between data source and evaluation measure in the survey.

\subsection{Findings}

The survey makes clear that the vast majority of papers that perform evaluation use either (1) synthetic data; or (2) empirical data combined with non-interventional measures (observational measures, structural measures, or visual inspection).  Our proposed ideal evaluation (empirical data and interventional measures) is used in only 7\% (6) of papers.  This raises an obvious question: Are the most commonly used evaluation techniques sufficient for determining whether algorithms for learning causal models will work effectively in realistic scenarios?  As we argue below, they are not.  

\section{The Case for Empirical Data}
\label{sec:case-empirical}
As already noted, nearly all causal modeling algorithms are ultimately designed for use outside of a laboratory, on real systems to infer useful causal knowledge about the world.  Despite this, evaluation of such algorithms often uses synthetic rather than empirical data.

\subsection{Limitations of Synthetic Data}

Researchers have developed several approaches to generating synthetic data.  The most common is to use a some form of directed graphical model.  In some cases, the structure of the model is designed to match the causal structure of a realistic system, either by manually specifying the structure or by learning it from empirical data.  Large-scale simulators designed for other reasons can also be used.  In some cases, simulators can be complex enough to generate data that is effectively equivalent to empirical data, though such simulations vary in quality.

Synthetic data is easy to collect, allows for straightforward comparison with ground truth, and facilitates systematic testing across a variety of data parameters.  Its popularity is evident---84\% (74) of surveyed papers used it in their evaluation, and 41\% (30/74) of those used only synthetic data.  However, using synthetic data for evaluation also has significant limitations.  These include:

\textit{Unquestioned assumptions}---Synthetic data tends to match the assumptions of the researcher running the study and any algorithms they have created. For example, a researcher developing an algorithm that outputs a DAG will be inclined to generate data from a DAG.

\textit{Unknown influences}---Even the best data generators can only include the influences already known to researchers.  Almost by definition, synthetic data generators cannot include any ``unknown unknowns'' that may influence the outputs of real-world systems.  While latent variables can be added, they are still defined and created by the researcher, limiting the realism of the data.

\textit{Lack of standardization}---Synthetic data is typically generated differently by each researcher, and this lack of standardization impedes comparison between studies.  

\textit{Researcher degrees-of-freedom}---Synthetic data is typically designed and parameterized by the researchers who created the algorithm being evaluated, giving them an enormous range of choices.  Such high ``researcher degrees-of-freedom'' \citep{simmons2011false} are a basic challenge to the validity of any study.

These factors significantly limit the external validity and realism of most synthetic data, making it insufficient as the sole source of data for evaluation.  Synthetic data is not without value---it can be a powerful way to assess features of an algorithm and test its performance under different conditions.  However, it typically falls short in providing insights into how the algorithm will perform on data from a real-world system.

\subsection{Benefits of Empirical Data}

Empirical data is almost always more difficult to collect than simulated data, and information on the effects of interventions is typically also much more difficult to obtain.  However, using empirical data has multiple benefits:

\textit{Realistic complexity}---Empirical data typically has a distribution that is more complex than simulated data.  That distribution is subject to realistic latent factors and measurement error.  This creates a learning task that is often significantly harder than synthetic data, but also more closely matches the challenges of real-world settings.  

\textit{Lower potential researcher bias}---Empirical data is typically not generated by the researcher who designed the algorithm being evaluated, and thus it is less subject to unintentional biases.  In addition, individual data sets are often shared across the community, creating standardization and comparability across studies.

\textit{Real-world demonstration}---The aim of research on algorithms for causal modeling is to have these algorithms used by others to infer causal models and reason about causal effects in real-world settings.  Practitioners considering use of these methods may be legitimately skeptical about their effectiveness until they see successful demonstrations of accurate causal inference on real-world data.

However, using empirical data poses challenges as well.  Because it is generally not collected by the person using it, some features of the data may not be fully understood, hindering correct interpretation.  Also, ground truth can be challenging to obtain, limiting evaluation to visual inspection or observational measures.  This is unsatisfying at best and misleading at worst, since, when evaluating without ground truth, it can be easy to see meaning where none exists or to imagine explanations for many possible conflicting outputs.  Despite these challenges, empirical data is still used frequently in practice; 67\% (61) of surveyed papers use empirical data, and 28\% (17/61) used only empirical data.

\subsection{Sources of Empirical Data}

Types of empirical data vary depending on the level of ground truth and the source of the ground truth.  Purely observational data is the most readily available and is used most often.  While this is rarely accompanied by full knowledge of the underlying structure, there are generally some dependencies that are known, either from common sense knowledge (such as temporal ordering) or from dependencies that have already been established by prior work. For a randomized controlled trial, the dependence between the measured treatment and outcome is generally taken as ground truth.  The same is true for cases in which multiple potential outcomes can be recorded for each unit.  This includes gene regulatory networks, flow cytometry analysis, and software systems, where essentially identical units can receive multiple treatments and thus produce multiple potential outcomes.

Because interventional measures and empirical data are used so infrequently, one might assume this is because such data sets are difficult to obtain.  This is partially true---there are significantly more observational data sets available than interventional data sets.  However, a growing community is producing data sets that provide interventional effects.  We describe some of them here.

The cause-effect pairs challenge \citep{mooij2016distinguishing} provides data that is empirical and, while interventional effects are not available, the direction of causality is known.  The 2016 Atlantic Causal Inference Conference Competition and subsequent competitions \citep{Dorie2019, hahn2019atlantic} created semi-synthetic data sets, producing synthetic treatment and outcome functions using covariates from a real-world system.  A similar approach was used by \citet{Shimoni2018} for the IBM Causal Inference Benchmarking Framework.  Flow cytometry data, measuring protein signaling pathways, is another common choice for interventional data \citep{sachs2005causal}.  \citet{Dixit2016} provide data on gene expression, collected using their proposed Perturb-Seq technique to perform gene deletion interventions.  There has also been work in partially randomized experiments, where a population is split into randomized and observational groups, creating parallel datasets for evaluation \citep{Shadish2008can}.  Other sources of interventional and empirical data include results of advertising campaigns \citep{Sun2015} and clinical studies \citep{McDonald1992}, as well as multiple challenges organized for machine learning conferences \citep{Guyon2008, Guyon2010}.  Domain specific simulations are another useful source of data. While technically synthetic, a sufficiently sophisticated simulation falls on a spectrum between purely synthetic and purely empirical data. They are often highly complex, are created by someone other than the researcher, and are created for a purpose other than evaluation, making them ideal for evaluation.  One popular simulation that is used for evaluation is the DREAM in silico data sets, since multiple combinations of single-gene interventions can be performed on identical networks \citep{Schaffter2011}.

We also introduce an additional source of empirical data where interventions are possible: large-scale software systems.  These systems have many desirable properties for the purposes of empirical evaluation: (1) They are pre-existing systems created by people other than the researchers for a purpose other than evaluating algorithms for causal modeling; (2) They produce non-deterministic experimental results due to latent variables and natural stochasticity; (3) System parameters provide natural treatment variables; and (4) Each experiment is recoverable, allowing the same experiment to be performed multiple times with different combinations of interventions.  Three such data sets are discussed in more detail in Section \ref{sec:example} and in the Supplementary Material.\footnote{These data sets are available for download at http://kdl.cs.umass.edu/data.}

\subsection{How Different are the Results?}

Readers may ask: In practice, what's the difference between using empirical data rather than synthetic data?  If that difference is small, then the substantial extra work involved in evaluation with  empirical data may not be worth the effort.

To begin addressing this question, we conducted a series of experiments using the interventional data from the software systems mentioned above. Specifically, we used a common approach for generating somewhat realistic synthetic data.  This approach uses an empirical data set to learn a causal model and then uses that model to generate synthetic data (and known ground truth) for model evaluation.  While the final data set is synthetic, its structure may better approximate the empirical system, rather than being entirely defined by the researcher, lending it more credibility.  We used this approach to generate synthetic data in the style of the three empirical data sets we generated from software systems.  Since we now have both empirical and synthetic data, each with ground truth, we can use causal modeling algorithms to construct a model for both of these data sets and compare the conclusions we would draw from each.

\begin{figure*}
\centering
\includegraphics[width=.59\pdfpagewidth]{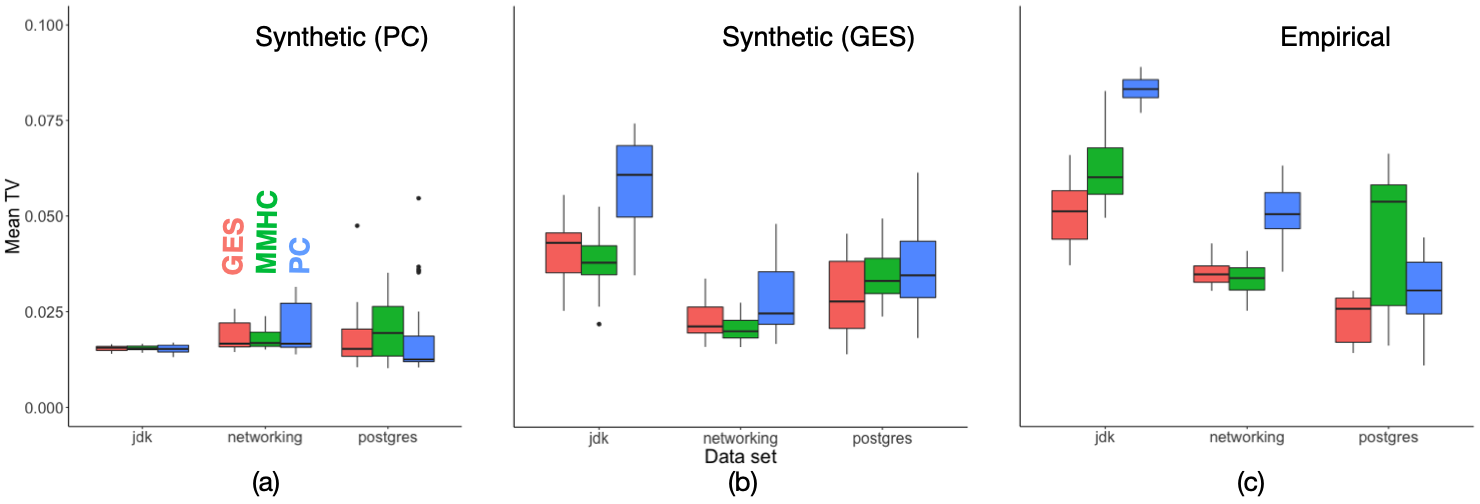}
\caption{Comparison of TVD on empirical data and synthetic data derived from empirical data.  \textbf{(a)} and \textbf{(b)}: synthetic data with structure obtained from PC or GES.  \textbf{(c)}: TVD on empirical data.}
\label{fig:empirical_results}
\end{figure*}

The synthetic data used was created by first choosing an initial causal modeling algorithm to create a ground truth model from the empirical data.  After learning a ground truth model with each of two algorithms that construct causal graphical models (PC and GES),\footnote{We reach similar conclusions based on the results for MMHC, which are reported in the Supplementary Material.} we generated synthetic data using the resulting models.  We then evaluated the same three algorithms on both the synthetic and empirical data.  Figure \ref{fig:empirical_results} shows how mean TVD varies for different causal modeling algorithms and different data sets.  The results shown are the mean TVD when evaluating PC, GES, and MMHC on two types of synthetic data sets (using the model as ground truth) and on the empirical data (using the known interventional effects).  There is significant variability between the two methods of generating the synthetic ground truth model from the empirical data (PC and GES), both in the mean TVD and in the relative ordering of the algorithms.  Comparing the synthetic and empirical results, some relative orderings of the algorithms are the same (e.g., network), but other orderings are significantly different (e.g., Postgres). These results suggest that algorithm performance cannot be expected to match between synthetic and empirical data, even when the synthetic data is created in a way that would be most expected to match aspects of the empirical data.


%

\section{The Case for Interventional Measures}
\label{sec:case-interventional}

Many algorithms are currently evaluated based on their ability to learn causal structure.  However, the actual desired task is almost never to model structure alone.  In practice, estimating the magnitude of interventional effects is vitally important, and an algorithm that cannot distinguish between strong and weak effects is severely limited in scope.  Despite this, the majority of current evaluations use observational or structural measures rather than measures of interventional effect.

\subsection{Limitations of Observational Measures}

Observational measures are widely used to evaluate algorithms for associational modeling, where the task of the algorithm is to discern statistical associations between two or more variables.  In such applications, the primary focus is effectively modeling the magnitude and form of statistical dependence, rather than explicitly learning causal dependence.  This highlights a severe and obvious limitation of observational measures:

\textit{Non-causal}---Observational measures are, by definition, not causal.  They measure the error of estimates of the outcome variable, but they do not measure that error under intervention.  They provide a sense of how well an algorithm has learned statistical dependence, but not how well it has learned causal dependence.  Despite this, observational measures are the \emph{only} evaluation used in 23\% (21/91) of papers surveyed.

\subsection{Limitations of Structural Measures}

Structural measures are easy to calculate, and they have a clear intuition. If an algorithm produces a causal structure and we know structural ground truth, it seems sensible to determine if the two structures match.  This has led to the widespread adoption of structural measures: 55\% (50) of surveyed papers used such measures, and 84\% (42/50) of those used only structural measures.  However, structural measures have several serious limitations:

\textit{Requires known structure}---Calculating structural measures requires a full ground-truth graph structure, which is only rarely available for empirical data.  

\textit{Constrains research directions}---The prevalence of structural measures may constrain research to algorithms that can be evaluated with these measures.  Algorithms that do not produce DAGs are less likely to be developed or favorably reviewed.  Since structural measures can only be used by algorithms that produce a directed graphical model as output, they implicitly assume that directed graphical models are capable of accurately representing any causal process being modeled, an unlikely assumption.

\textit{Oblivious to magnitude and type of dependence}---Structural measures, by design, do not account for different magnitudes of dependence, so an error in an edge with a strong effect incurs the same penalty as an error in an edge with a very weak effect.  In addition, structural measures are only able to measure \emph{which} variables in a causal model change as the result of an intervention.  In many cases, it is also necessary to determine \emph{how much} or \emph{in what way} a given target quantity will change with respect to an intervention.

\textit{Oblivious to likely treatments and outcomes}---In most cases, structural measures do not consider where an edge is located in the overall structure of the DAG, so an edge with many downstream effects is treated the same as a less central edge.

\subsection{Benefits of Interventional Measures}

In contrast to observational and structural measures, interventional measures have strong advantages:

\textit{Correspondence to actual use}---Interventional measures evaluate how well the model estimates interventional effects, which aligns more closely with the eventual use of nearly all causal models.  For example, a directed acyclic graph is not the ultimate artifact of interest for most applications---DAGs are a representation that facilitates estimation of interventional effects \citep{spirtes2000causation, pearl2009causality}. Thus, it seems natural to define an evaluation measure in terms of interventional effects rather than graphical structure.  

\textit{Weighting of different errors}---While most structural measures penalize each edge misorientation equally, interventional measures penalize misorientation errors proportionally to their effect on the estimation of interventional effect.



\subsection{How Different are the Results?}

Interventional measures are intended to capture something different than structural measures, but they are ultimately affected by the structure of the learned model, and we would expect structural errors to lead to interventional errors.  Of course, interventional and structural measures are equal when structure and parameterizations are perfect, but they can differ significantly when the learned structure is only approximately correct (which is almost always the case).  To assess the extent to which interventional measures capture different information than structural measures in such cases, we ran experiments using synthetic data.  This allowed us to produce data where we could calculate both structural measures and interventional measures, since we have the full parameterized ground truth model to compare against.

\setlength{\abovecaptionskip}{-3pt}
\begin{wrapfigure}[12]{r}{0.5\textwidth}
\vspace{-2pt}
\includegraphics[width=0.47\textwidth]{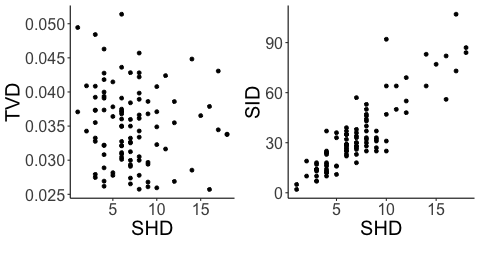}
\caption{Structural and interventional measures compared on synthetic data with GES.}
\label{fig:ges_synthetic}
\end{wrapfigure}

For these experiments, we produced data from random DAG structures with conditional probability models drawn from a Dirichlet distribution.  We generated 5000 instances, applied a causal modeling algorithm, and calculated various evaluation measures.  Figure \ref{fig:ges_synthetic} shows the results for GES.   SHD and SID are clearly strongly correlated, suggesting that both structural measures ultimately produce similar quality measures of the algorithm.  However, SHD and TVD are only very weakly correlated, with many models scoring highly with one measure and poorly with the other.  At least in this case, the interventional measure (TVD) appears to capture substantially different information than that of a structural measure (SHD).  Results for PC and MMHC are reported in the Supplementary Material.

\section{Example of an Evaluation}
\label{sec:example}

\begin{figure*}[ht]
\vspace{-11pt}
    \centering
    \includegraphics[width=.6\pdfpagewidth]{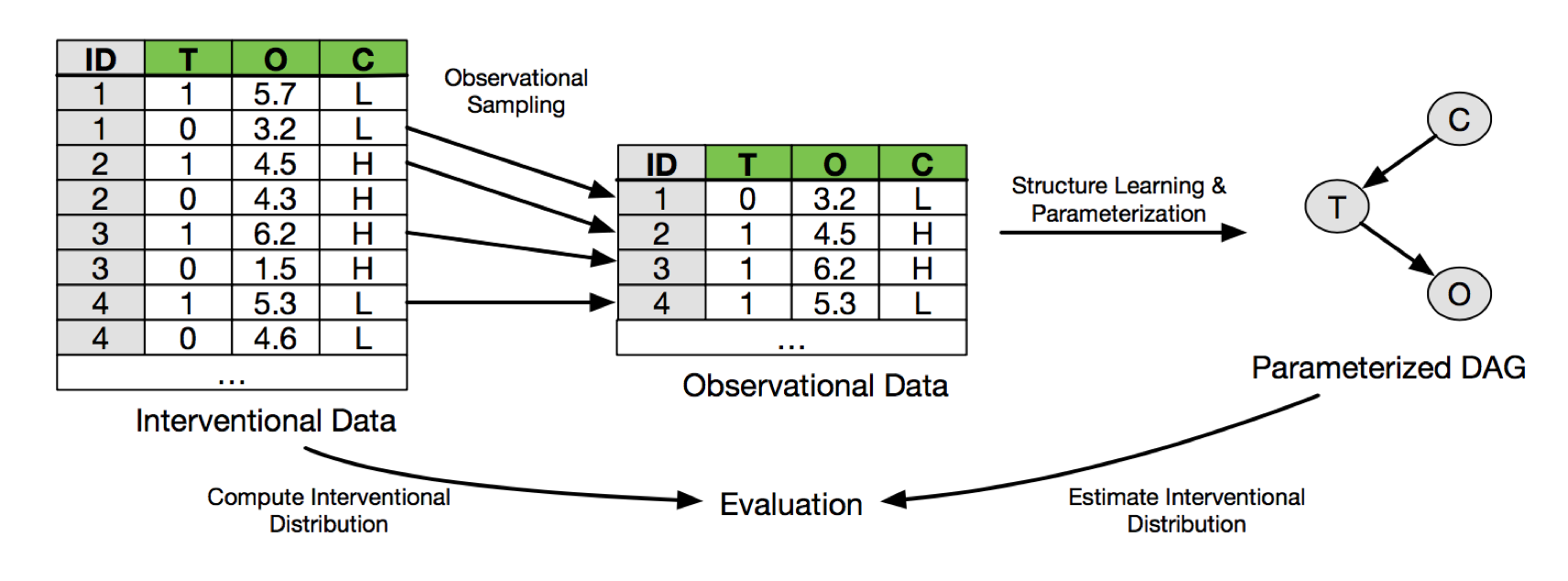}
    \caption{A diagram of one way to evaluate a causal modeling algorithm}
    \label{fig:example-flow}
\end{figure*}

To further explain what we mean by empirical data and interventional measures, we describe one example of this type of evaluation, shown schematically in Figure \ref{fig:example-flow}.  This example demonstrates one way that an evaluation with empirical data and interventional measures could be performed, though many other techniques are possible, depending on the algorithm, data source, evaluation measure, and the research question under consideration.  In our example, we evaluate the PC algorithm \citep{spirtes2000causation}, Greedy Equivalence Search (GES) \citep{chickering2003optimal}, and MMHC \citep{tsamardinos2006maxmin} by measuring \textit{total variation distance} (an interventional measure defined later) on a data set produced by experimentation with a large-scale software system.  

The most obvious way to evaluate how well an algorithm can learn causal models from real-world data is to compare the model's estimates to empirical data drawn from a system in which we can perform multiple interventions on the same units, giving us full interventional data in which we can assess every potential outcome for each unit.  Large-scale software systems allow for this type of intervention because they let us run the same experiments multiple times under different conditions (e.g., different settings of key system parameters).  An example of this is a Postgres database, where we can run the same queries with different settings of key configuration parameters.  In this context, each query corresponds to a unit, a set of configuration parameters correspond to treatment, and variables such as runtime correspond to outcomes.  Details about this data can be found in the Supplementary Material.

Many algorithms for causal modeling are designed to run on observational data, in which only a single, non-randomized treatment assignment is observed for each unit.  In the absence of an observational data set that matches our interventional data, we can create an observational-style data set by sub-sampling the full interventional data in a non-random manner.  To do this, we select a single treatment assignment for each query.  Selecting treatment at random is equivalent to a randomized controlled trial.  In most observational contexts, however, treatment assignment would be based on covariates of the unit.  For example, a database administrator might choose the configuration parameters based on features of each query.  We use a similar process to create observational data by using a measured covariate of the query to probabilistically assign treatment.

\pagebreak

Given such an observational data set, we can apply a causal modeling algorithm and learn a causal model.  A fully parameterized model can produce an estimated interventional distribution $\hat{P}$ by applying the \emph{do}-calculus~\citep{galles1995testing}.  Under this framework, causal quantities take the form of probability queries with $do$ operators, for instance $P(O|do(T=1))$.  We can also estimate the actual interventional distribution $P=P(O = o|do(T = t))$ for any outcome $o$ and treatment $t$, because we can measure the effects of both values of treatment for each query in our data set.

We then can use an interventional measure to compare the true interventional distribution $P$ to the estimated distribution $\hat{P}$.  One example of an interventional measure is total variation distance (TVD) ~\citep{lin1991divergence}, which measures the distance between two probability distributions. For discrete outcomes $O$, the quality of an estimated interventional distribution relative to a known distribution under TVD is straightforward to compute:

\small
\setlength{\abovedisplayskip}{-5pt}
\begin{align*}
\label{eq:total-variation}
TV_{P, \hat{P}, T=t}(O) = \frac{1}{2} \sum_{o \in \Omega(O) } \big| &P \left( O = o|do(T=t) \right) -  \hat{P}\left(O=o|do(T=t) \right) \big|,
\end{align*}
\normalsize
\vspace{-8pt}

where $\Omega(O)$ is the domain of $O$.  This gives us a numerical measure of how well the estimated interventional estimates match the ground truth.  A single TVD value is computed for each causal effect, which can then be aggregated for comparison.  Results of this evaluation on the software data is shown in Figure \ref{fig:empirical_results}c.  For these datasets, we can conclude that GES has the best overall performance.

\section{Conclusion}
\label{sec:conclusions}
Evaluation is a key mechanism that determines how algorithms are viewed within the community, what research directions are pursued next, and whether our research has broader impacts outside the community.  Our current evaluation techniques aim too low, and they fail to evaluate the full range of questions that our research goals imply.

We are not the first to point out the need for more robust evaluation techniques.  Some of the datasets we discuss were created in response to recognition that better evaluation was necessary \citep{Dorie2019, Shimoni2018, mooij2016distinguishing}. In addition, prior work has examined the importance of testing the generalizability of causal inferences drawn from observational data \citep{zhao2019, keane2007exploring} and comparing causal effects drawn from observational and experimental data \citep{cook2008three, Eckles7316, eckles2017bias, gordon2019comparison}.  However, despite this, as our survey shows, empirical evaluation with interventional measures is rarely used by computer science researchers.

We acknowledge that, while the evaluation techniques we advocate are applicable to a wide range of algorithms, data sets may not be available for every task.  The diverse tasks of causal modeling algorithms make it difficult to recommend a single data set and evaluation measure to evaluate every algorithm.  However, the data sets and measures that are most commonly used are largely insufficient.  The community would benefit if more data sets with interventional effects were created and made available for public use, allowing for a breadth of evaluation options.

We do not advocate abandoning synthetic data and structural measures.  Both have many uses for evaluating algorithm performance and can be indispensable scientific tools.  However, they are insufficient on their own.  Instead, they should be viewed as a first step in evaluation. If we want causal modeling algorithms to be adopted outside our research community, we need demonstrations of their utility outside of a laboratory setting.  If we do not evaluate on empirical data, we cannot be certain our algorithms will perform well on real data, and if we do not evaluate with interventional measures, we cannot be certain that the causal effects the algorithm infers will translate to actual, substantial causal effects.  Expanding our routine evaluations will substantially improve the credibility and comparability of results, the external validity and trustworthiness of algorithms, and the efficiency with which we conduct our research.



\subsubsection*{Acknowledgments}
This material is based upon work supported by the United States Air Force under Contract No, FA8750-17-C-0120.  Any opinions, findings and conclusions or recommendations expressed in this material are those of the author(s) and do not necessarily reflect the views of the United States Air Force.

\bibliography{causal-eval.bib}
\bibliographystyle{plainnat}

\end{document}


\maketitle
\section{Additional Details on Software Data}
We introduce a source of empirical data where interventions are possible: large-scale software systems.  We performed experiments on three large computational systems: Postgres, the Java Development Kit, and HTTP processing.  These systems have many desirable properties for the purposes of empirical evaluation: (1) They are pre-existing systems created by people other than the researchers for a purpose other than evaluating algorithms for causal discovery; (2) They produce non-deterministic experimental results due to latent variables and natural stochasticity; (3) System parameters provide natural treatment variables; and (4) Each experiment is recoverable, allowing the same experiment to be performed multiple times with different combinations of interventions.

Within each computational system, we measure three classes of variables: outcomes, treatments, and subject covariates.  Here, outcomes are measurements of the result of a computational process, treatments correspond to system configurations and are selected such that they could plausibly induce changes in outcomes, and subject covariates logically exist prior to treatment and are invariant with respect to treatment.  Using these variables, we can apply all combinations of treatments to all subjects, and we can use these results to estimate actual interventional distributions for the effects of each treatment variable on each outcome variable.  We can also then sub-sample these experimental data sets in a manner which simulates observational bias to produce observational-style data sets, allowing us to evaluate an algorithm's performance on pseudo-observational data and evaluate it using actual interventional effects.  These data sets will be made available after publication.

We had a number of goals in mind when gathering data from our real domains:
\begin{itemize}
    \item \textbf{Causal Sufficiency:} The algorithms we studied require that no pair of variables in the model are both caused by a latent variable. We can guarantee this is true for pairs of treatments and outcomes (since treatments have no parents in the original data set), but needed to employ domain knowledge to limit sources of causal sufficiency violations with regard to other pairs of variables.
    \item \textbf{Acyclicity:} Each of the systems can be described by a ``single-shot'' computational process which starts and finishes without the possibility for feedback.
    \item \textbf{Instance Independence:} We took efforts to ensure that each execution of the computational process was independent of previous executions. In most cases, this required clearing caches and resetting other aspects of system state.
    \item \textbf{Plausible Dependence:} We selected variables that we believed would be causally related. 
\end{itemize}
Each domain is characterized by three classes of variables: subject covariates, treatments, and outcomes. Under the factorial experiment design, outcomes were measured for every combination of subjects and treatments. This yields a data set with many records for the same subject, as in the example in Table~\ref{tbl:factorial-experiment-example}. To permit greater opportunities for observational sampling, we performed multiple trials of each factorial experiment.
\begin{table}[]
\centering
\begin{tabular}{r | r | r | r}
Subject ID & Covariate & Treatment & Outcome \\
\hline
1 & A & 0 & 1.33  \\
1 & A & 1 & 0.96 \\
2 & B & 0 & 1.89 \\
2 & B & 1 & 0.54 \\
3 & A & 0 & 1.02  \\
3 & A & 1 & 0.99 \\
4 & A & 0 & 1.35 \\
4 & A & 1 & 1.12 
\end{tabular}
\caption{An Example of a Factorial Experiment with Four Subjects and a Binary Treatment}
\label{tbl:factorial-experiment-example}
\end{table}
Given the difficulty associated with modeling highly complicated outcomes such as runtime, we employed a normalization scheme for each data set, dividing outcome values by a ``baseline'' value---the median control-case outcome value. Thus, we ultimately recorded outcomes which represent a deviation from this baseline. In this regard, our experimental results resemble a within-subjects design \cite{greenwald1976within}, although without many of the pitfalls that plague experiments on humans, such as non-independence of outcome measurements.
In the original data from each domain, subject covariates are either discrete, continuous, or binary; treatments are binary; and outcomes are continuous.
We converted each of the variables to a discrete representation to make parametrization and inference more robust.
\subsection{Java Development Kit}
\begin{figure}
\centering
\includegraphics[width=.75\linewidth]{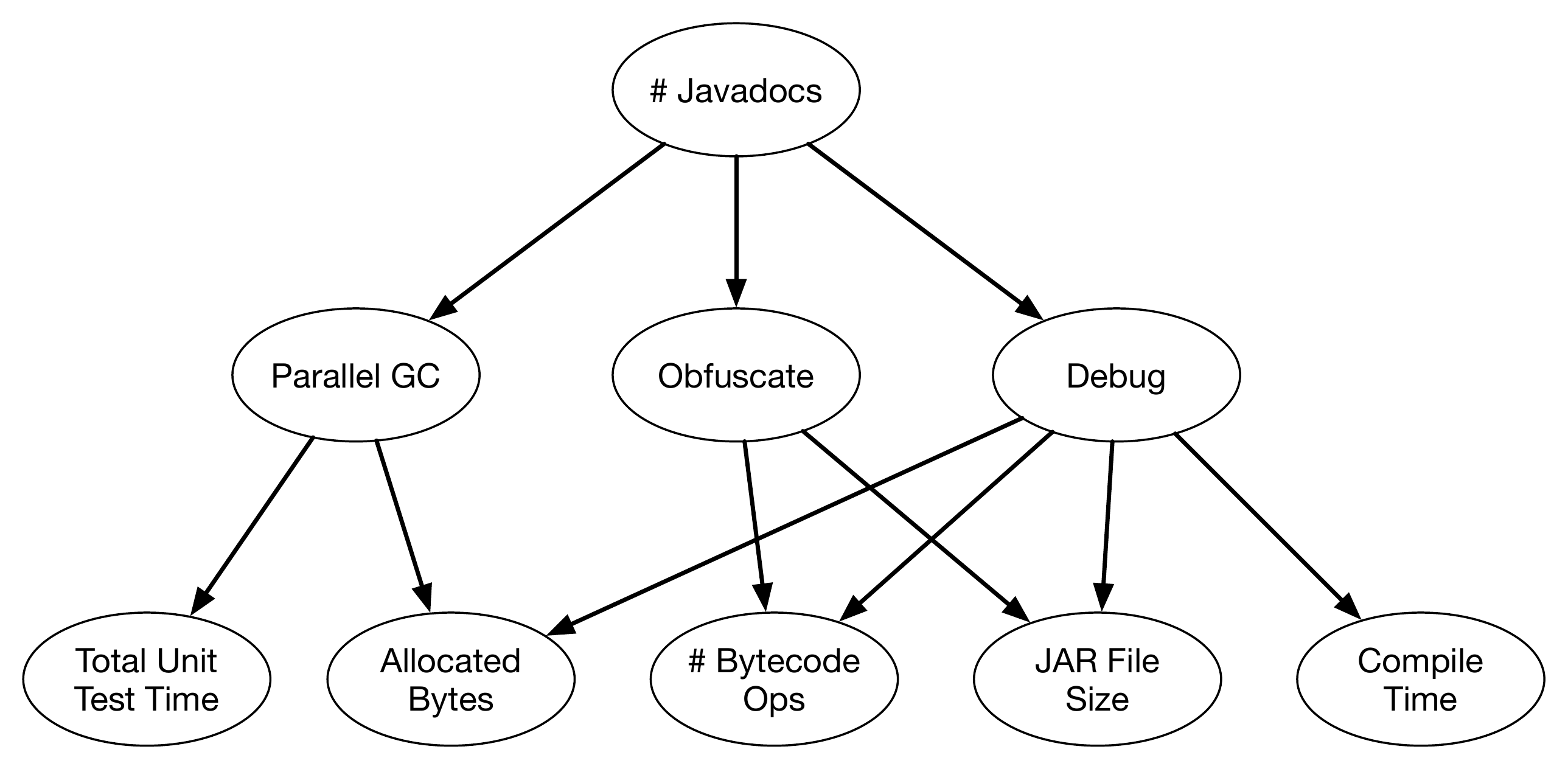}
\caption{Consistent Model for the JDK Domain}
\label{fig:jdk-consistent}
\end{figure}
Our experiments on the Java Development Kit (version \texttt{1.7.0\_60}) used 2,500 Java projects obtained from GitHub as the subjects under study. We retrieved only projects which use the Maven build tool to facilitate automated compilation and execution.
Additionally, we constrained our search to include only projects which had unit tests.
This may introduce selection bias in our data collection processes, but this is acceptable. It is not important that our conclusions generalize to some population of computational systems, only that there are causal dependencies which hold on the sub-population under investigation.
Of those, 473 compiled and ran without intervention. This group yielded a total of 7,568 subject-treatment combinations.
For each combination, we compile and execute the unit tests of the Java project.
In order to obtain full state recovery between each trial, any compiled project files were cleared between executions.
Thirty-five CPU days were required to collect this data using several Amazon EC2 instances.
\subsubsection{Treatments}
\begin{itemize}
    \item \textbf{Aggressive Compiler Optimization}: Disabling this option (enabled by default) prevents some compiler optimizations from running, potentially slowing down execution time but perhaps reducing compilation time. This option is disabled with the \texttt{javac} option \texttt{-XX:+AggressiveOpts}.
    \item \textbf{Emission of Debugging Symbols}: Debugging symbols are used to provide a map through the compiled source code that can be used for interactive debugging and diagnostics. Inclusion of these symbols may require some time during the compilation phase, increase the size of the compiled program, and could possibly impact runtime. This corresponds to the \texttt{-g} flag of \texttt{javac}.
    \item \textbf{Garbage Collection Methodology}: The Java Development Kit supports several garbage collection schemes. Two were considered: parallel and serial. These schemes are activated with the \texttt{-XX:-UseParallelGC} or \texttt{-XX:-UseSerialGC} arguments.
    \item \textbf{Code Obfuscation}: Several third-party tools are capable of obfuscating compiled code, making reverse-engineering difficult. This process could also affect the size of the compiled project files. The yGuard\footnote{http://www.yworks.com/en/products\_yguard\_about.html} tool was used for this purpose.
\end{itemize}
\subsubsection{Outcomes}
\begin{itemize}
    \item \textbf{Number of Bytecode Instructions}: Before execution, Java code is compiled to an intermediate language referred to as bytecode. We measured the number of atomic instructions, or operations, in this compiled code to form this outcome using a custom-built bytecode analysis tool based on Javassist\footnote{http://www.csg.ci.i.u-tokyo.ac.jp/~chiba/javassist/}.
    \item \textbf{Total Unit Test Time}: Each project we gathered contains one or more unit tests. To capture the runtime of the full unit test workload, we computed the sum of runtimes of all unit tests for a given project.
    \item \textbf{Allocated Bytes}: The Java Virtual Machine supports a profiling option (\texttt{-agentlib:hprof=heap=sites}) which can be used to track heap statistics throughout a program's execution. We utilized this feature to obtain the total number of bytes allocated during unit test execution.
    \item \textbf{Compiled Code Size}: Java programs are often packaged in an format known as a JAR (Java ARchive). To characterize the size of the compiled code, we recorded the size in bytes of the associated JAR file.
    \item \textbf{Compilation Time}: In order to execute unit tests, the entire project needs to be compiled. This outcome represents the time used to convert all source files to their bytecode equivalents.
\end{itemize}
\subsubsection{Subject Covariates}
All subject covariates were obtained using the \mbox{JavaNCSS} tool\footnote{http://javancss.codehaus.org/}.
\begin{itemize}
    \item \textbf{\# NCSS (non-comment source statements) in Project Source}: This covariate is highly predictive of compiled code size. Conceivably, in observational settings, large projects could also be associated with more liberal use of advanced compilation settings and tools, such as a code obfuscator.
    \item \textbf{\# NCSS, Functions, and Classes in Unit Test Source}: These covariates are somewhat representative of the unit test workload. Projects with many lengthy unit tests may also have longer total unit test runtime.
    \item \textbf{\# ``Javadoc'' comments in Unit Test Source}: This covariate could be indicative of code quality. Well-commented code is perhaps more likely to be found in high-quality projects. This code may be more likely to be used in production environments, and thus could be less likely to be observed with debugging symbols. This feature is used in the treatment-biasing procedure for construction of observational data sets.
\end{itemize}

\subsection{Postgres}
\begin{figure}
\centering
\includegraphics[width=.75\linewidth]{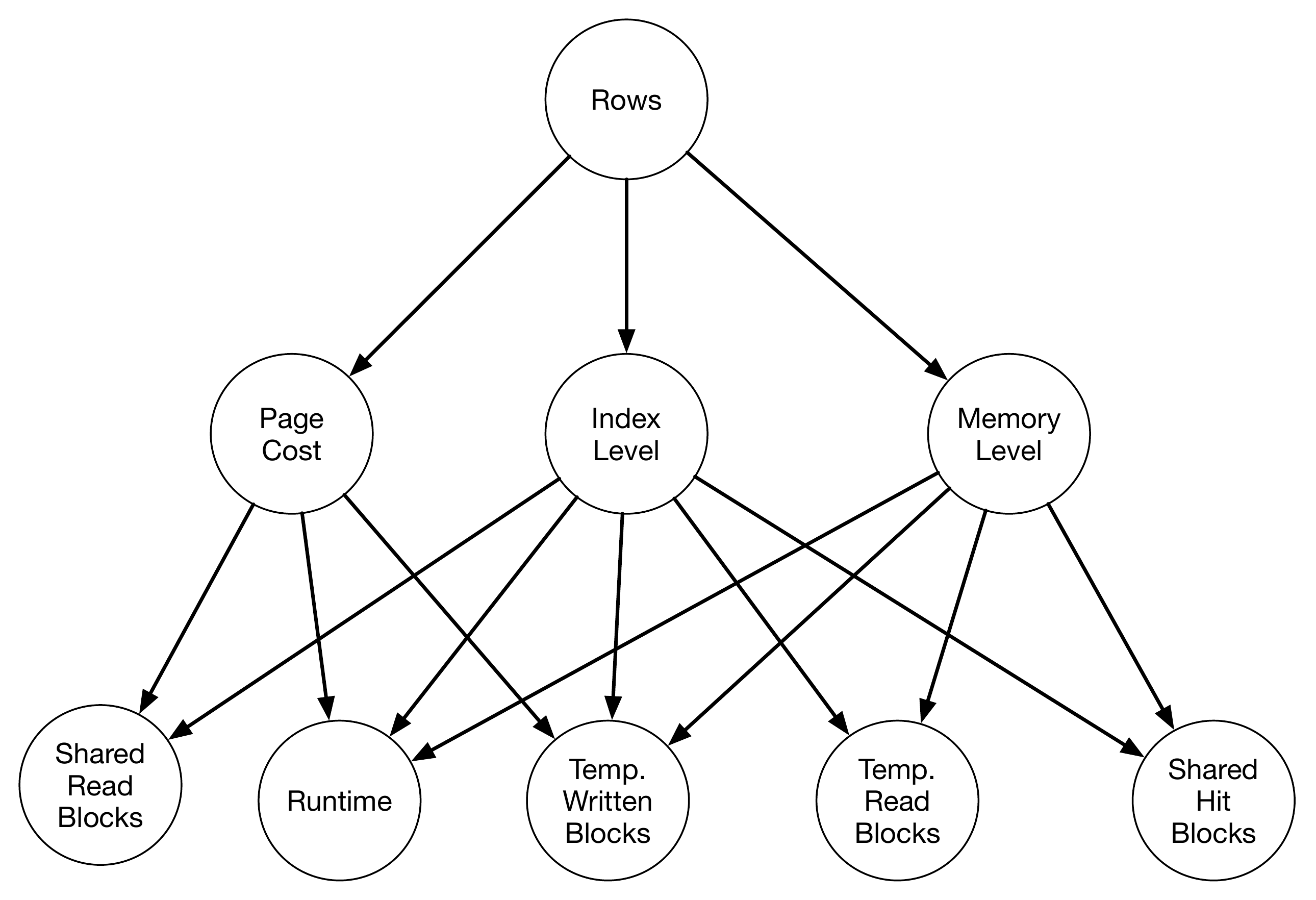}
\caption{Consistent Model for the Postgres Domain}
\label{fig:postgres-consistent}
\end{figure}
Consistent with a data warehousing scenario, we employ a fixed database for our Postgres (version 9.2.2) experiments: a sample of the data from Stack Overflow, drawn from the Stack Exchange Data Explorer\footnote{http://data.stackexchange.com/}.
The data explorer also houses many user-generated queries.
We collected 29,375 of the most popular queries to use as subjects for this study.
Stack Exchange's data warehouse uses Microsoft SQL Server, which does not completely overlap with Postgres in supported features and syntax.
Some queries use only ANSI-compliant syntax and run successfully on either SQL Server or Postgres.
To obtain as large a set of subjects as possible, we employed a semantics-preserving query rewriting scheme to adapt queries into Postgres-compliant syntax wherever possible.
This yielded a set of 11,252 user-generated queries which executed successfully within Postgres for a total of 90,016 subject-treatment combinations.
In order to recover system state between trials, the shared memory setting (specifying how much main memory Postgres can use for caching) was set to 128 kilobytes, limiting caching significantly. Any queries which required more than 30 seconds to execute were marked as ``failures'' in order to prevent long-running queries from holding up other queries, which typically required one second to execute. As with the JDK data set, this may induce sampling bias, but we are not aiming for our experimental findings to generalize to the broader population of database queries.
\subsubsection{Treatments}
\begin{itemize}
    \item \textbf{Indexing}: A common administration task is to identify indices that can be used to accelerate lookup of commonly-referenced columns with a particular value or falling within a range. For our experiments, we employed two indexing settings: no indexing, and indexing on primary key/foreign key fields. Domain knowledge suggests that that the latter approach would dramatically reduce runtime of some queries. In all cases, the default B-tree index was employed.
    \item \textbf{Page Cost Estimates}: In order to determine if an index should be used, the database employs estimates of the relative cost of sequentially accessing disk pages and randomly accessing disk pages. 
We utilized two extremes for this setting: one scheme in which random page access is estimated to be fast, relative to the sequential page access, and one scheme in which the opposite relation holds. The corresponding database settings we adjusted were \texttt{random\_page\_cost} and \texttt{seq\_page\_cost}.
    \item \textbf{Working Memory Allocation}: The database engine can make use of fast random-access memory, if available, to store intermediate query results. The amount of working memory that is allocated to the system can be controlled with a configuration option. For our investigation, we employed a low-memory setting and a high-memory setting, with background knowledge suggesting that the latter would result in faster-executing queries. This treatment was instrumented with the \texttt{work\_mem} and \texttt{temp\_buffers} options.
\end{itemize}
\subsubsection{Outcomes}
\begin{itemize}
    \item \textbf{Blocks Read from Shared and Temporary Memory}: These two outcomes identify the number of blocks, or memory regions, that were read during query execution. Shared memory is persistent (disk) and is accessed during normal table-retrieval procedures. Temporary memory is volatile (main memory) and is used for staging ordering or joining operations.
    \item \textbf{Blocks Hit in Shared Memory Cache}: This outcome represents the number of memory reads that were to be performed against shared memory, but were identified instead in a main memory cache.
    \item \textbf{Runtime}: The total time to execute the query.
\end{itemize}
\subsubsection{Subject Covariates}
\begin{itemize}
    \item \textbf{Year of Query Creation}: The year that the query was entered on the Stack Exchange data explorer.
    \item \textbf{Number of Referenced Tables}: The number of distinct tables that are referenced in the query.
    \item \textbf{Total Number of Rows in Referenced Tables}: The sum of cardinalities of tables referenced in the query.
    \item \textbf{Number of Join Operators}: The number of join operators employed in the query, requiring merging data from two tables.
    \item \textbf{Number of Grouping Operators}: The number of grouping operators employed in the query, requiring reduction and possibly summarization of the data.
    \item \textbf{Number of Other Queries Created by the Same User}: The total number of queries that the Stack Exchange user has created.
    \item \textbf{Length of the Query in Characters}: The length of the query after application of relevant rewrite rules.
    \item \textbf{Number of Rows Retrieved}: The number of rows that are returned by the query. Logically, this value exists prior to application of any treatment and is invariant with respect to treatment (since the database is fixed), even though we can only measure it after query execution.
\end{itemize}
\subsection{Hypertext Transfer Protocol}
\begin{figure}
\centering
\includegraphics[width=.75\linewidth]{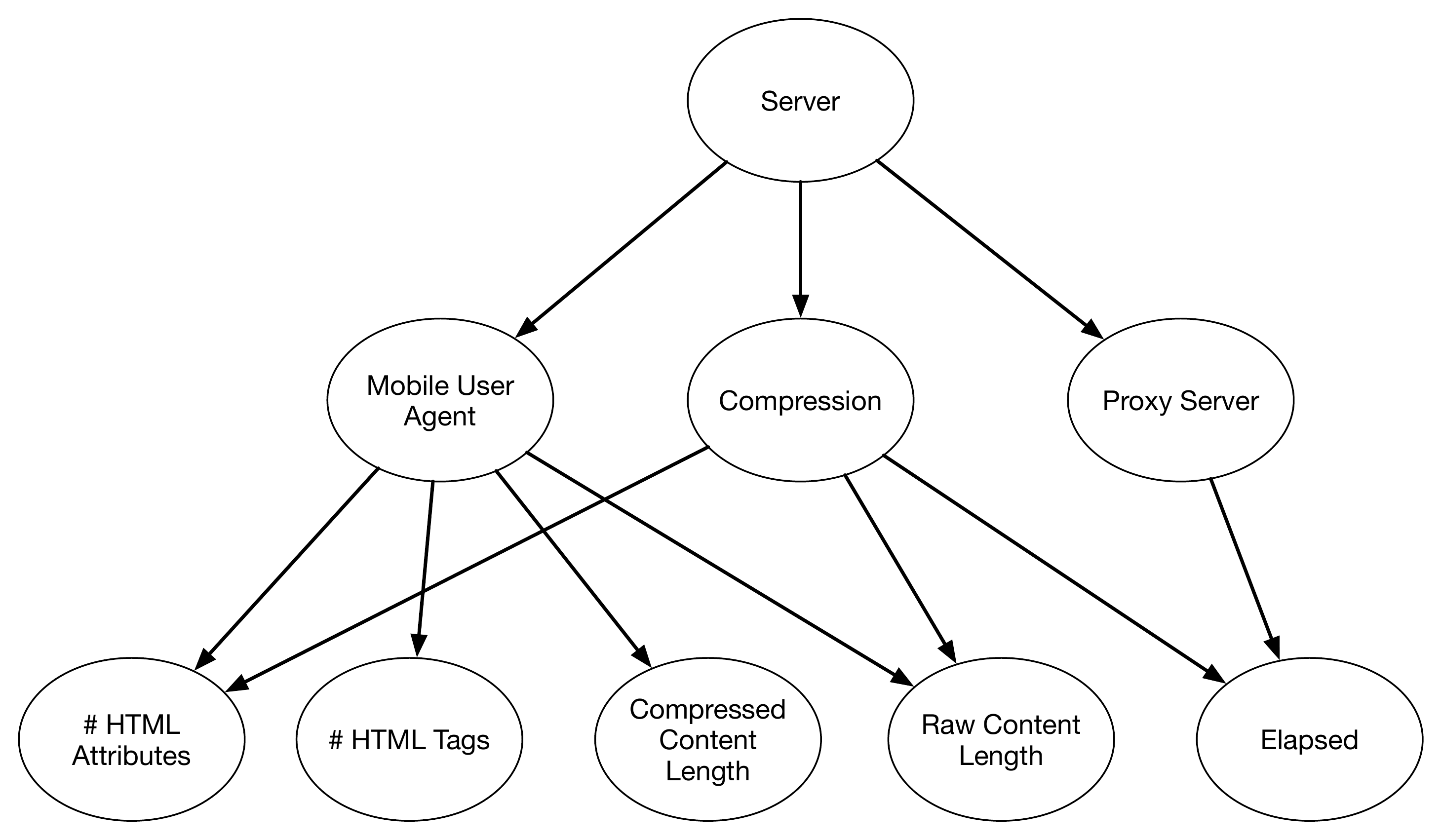}
\caption{Consistent Model for the HTTP Domain}
\label{fig:http-consistent}
\end{figure}
For our experiment on HTTP \& networking infrastructure, we used requests to specific web sites as subjects. We identified a number of target sites through a breadth-first web crawl initiated at \texttt{dmoz.org}. We ended the crawl after retrieving 5,472 sites. For 4,350 of those sites, we were able to issue successful web requests with all combinations treatments, yielding 34,800 subject-treatment combinations. We employed numerous techniques to ensure that content would not be cached, which could induce carryover across treatment regimes.
\subsubsection{Treatments}
\begin{itemize}
    \item \textbf{Use of a Mobile User Agent}: Web browsers supply a \emph{user agent} to identify themselves to the web servers that they request pages from. Some sites have different versions for mobile applications. We artificially adjusted the user agent from a standard user agent to a mobile user agent to explore this phenomena. This is accomplished with the HTTP \texttt{User-Agent} header.
    \item \textbf{Proxy Server}: Web requests can be routed through a \emph{proxy}, a server which issues web requests on behalf of a client. The additional time required to route the request to and from the proxy server can increase the elapsed time of the request. Our experiments were executed with Amazon EC2. Our ``client'' computers were making web requests from the east cost of the United States, and a proxy server was set up on the west coast.
    \item \textbf{Compression}: Applications can use the HTTP protocol to request that content be delivered with or without compression, possibly reducing the cross-network transmission time. In one compression configuration, the client requests \texttt{identity} compression, indicating that the content should be transmitted at face value. In another compression scheme, the client requests \texttt{gzip}, a common and effective scheme for HTTP content compression.
\end{itemize}
\subsubsection{Outcomes}
\begin{itemize}
    \item \textbf{\# of HTML Attributes and Tags}: These two outcomes describe the logical structure of the page. They may vary with respect to ``mobile user agent''.
    \item \textbf{Elapsed Time}: The time between issuance of the request and receipt of a response. This could be affected by network characteristics, which are determined in part by the time at which the request is issued and whether a proxy server is employed. Requests containing smaller payloads (influenced by compression) may also be faster to service.
    \item \textbf{Decompressed and Raw Content Length}: Two outcomes representing the size of a web page before and after content decompression, if applicable.
\end{itemize}
\subsubsection{Subject Covariates}
Only one subject covariate was identified for the HTTP domain, the web server reported via the \texttt{Server} header. This variable was coarsened into a version with 7 levels: Apache/2, Other Apache, Microsoft-IIS, nginx, Other,  and Unknown.
\section{Identifying Consistent DAGs}
To identify DAGs that can consistently estimate the all interventional distributions $P(O|do(T))$, we need to ensure that (1) the parent set of $T$ is a valid adjustment set with respect to $O$, and (2) if $T$ has a causal effect on $O$, there is a chain connecting $T$ and $O$ in the DAG model. The first condition is straightforward to satisfy since we know the only parent of any treatment to be the covariate used to introduce observational bias. The second condition requires identification of which pairs of treatments and outcomes are causally related. These \emph{d}-connection properties were identified for each domain using the full interventional data set using the Friedman test for blocked difference in means, allowing for correction of subject variability~\cite{friedman1937use}. An edge was introduced between any  causally related pair to satisfy condition (2). Then, ground truth interventional distributions $P(O|do(T=t))$ were produced by applying the do-Calculus model adjustment rules, and answering probability queries $P(P|T=t)$ on the resulting model using belief propagation.
\section{Pseudo-Observational Configurations}
We can transform the factorial experiments on our real domains into pseudo-observational data by sub-sampling the experimental data in a way that is correlated with a ``subject covariate''. This mirrors the process of treatment self-selection common to observational data. This transformation is outlined in Algorithm~\ref{alg:passive-treatment}.
\begin{algorithm}
\DontPrintSemicolon
\KwIn{Interventional data set $I$, biasing strength $\beta \geq 0$, biasing covariate $C$}
\KwOut{Observationally biased data set $O$, $|O| = nd$}
$l \gets $ The number of distinct values of $C$ \;
\ForEach{Subject $e \in I$}
{
    Let $C_e \in \{1..l\}$ represent the $C$ value of subject $e$ \;
    $Assign \gets \{\}$ \;
    \ForEach{Treatment $T_j$}
    {
        $s_{ej} \gets \begin{cases} 
                1 & \text{if $C_e \times j$ is even} \\
                -1 & \text{if $C_e \times j$ is odd} \\
        \end{cases}$ \;
        $p \gets \text{logit}^{-1}(s_{ej} \beta)$\;
        $t_j \gets $ Bernoulli$(p)$ \;
        $Assign \gets Assign \cup \{ T_j = t_j \}$ \;
    }
    $M \gets $ Record in $I$ corresponding to $(e, Assign)$ \;
    $O \gets O \cup M$ \;
}
\caption{Logistic Sampling of Observational Data}
\label{alg:passive-treatment}
\end{algorithm}

\section{Limitations of Empirical Data}
In the paper, we discuss popular sources of empirical data that is suitable for evaluation.  These data sets differ significantly in many ways, including level of realism and data quality, and they each have different benefits and limitations.

The cause-effect pairs challenge \citep{mooij2016distinguishing} provides observational data on pairs of variables where the direction of causality is known from domain knowledge.  This data set is useful for evaluating bivariate orientation algorithms, but the lack of any additional measured covariates limits its utility for evaluating multivariate structure learning algorithms.

The 2016 Atlantic Causal Inference Conference Competition data \citep{Dorie2019} and the IBM Causal Inference Benchmarking Framework \cite{Shimoni2018} use covariates taken from a real-world data set, allowing for potentially complicated interactions between them.  Treatment and outcome functions were then generated synthetically, using a variety of data generating processes to allow for the construction of many data sets with different features.  This allows algorithms to be tested on many data sets, providing a more robust evaluation.  However, the need to construct synthetic treatment and outcome functions limits the level of realism.

The software data we collected contains measurements of covariates, treatments, and outcomes from three real-world systems.  While the treatment function is generated synthetically, the outcome function is not, lending the ground truth causal effects from treatment to outcome a high degree of realism.  However, as with the above ACIC and IBM data sets, the treatment function still needs to be synthetically defined.

The flow cytometry data provided by \citet{sachs2005causal} contains measurements of protein signaling pathways, where multiple activating and inhibitory interventions were performed.  However, the ground truth is not clearly obtainable and most analysis using this dataset relies on structural measures.

Partially randomized experiments, where a population is split into randomized and an observational groups, are another useful source of empirical data \citep{Shadish2008can}.  The collection of randomized data drawn from the same base population as observational data creates a convenient ground truth for causal effect estimation.  However, due the nature of these experiments, they require careful experimental design to make sure the populations are equvalient and the treatments are correctly assigned and measured.

The DREAM in silico data sets \citep{Schaffter2011} are taken from a sophisticated simulation derived from multiple known gene regulatory network structures, which, while non-empirical, is intended to be complex enough to approximate empirical data.  However, realism is limited due to the use of a simulator.

\section{Additional Experiments}
\begin{figure}
\centering
\includegraphics[width=0.8\linewidth]{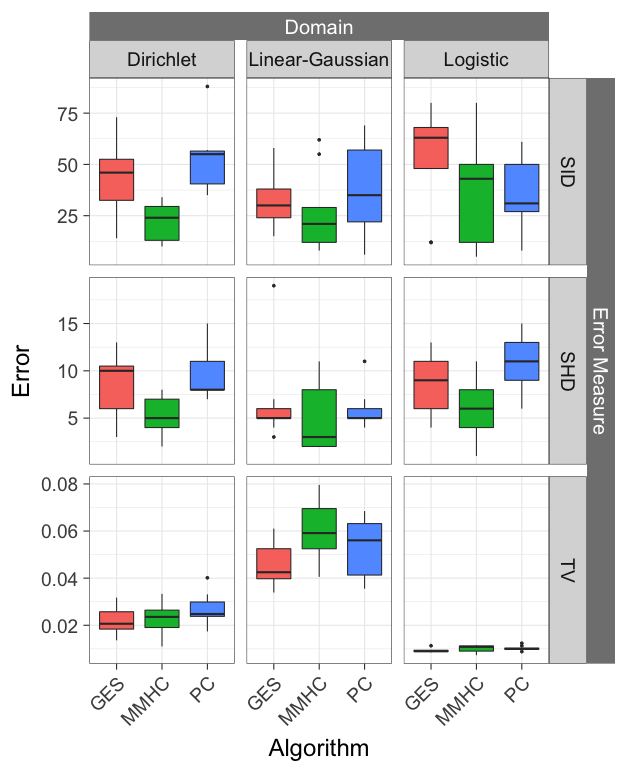}
\caption{Relative Performance of Causal Discovery Algorithms on Synthetic Data Sets}
\label{fig:synthetic_comparison}
\end{figure}

In the paper, we provided experiments that demonstrate that TVD and structural measures provide different information and that information is relevant for over and under specification.  To expand on these results, we performed an additional experiment to evaluate if different types of measures would lead to different conclusions about the relative performance of causal modeling algorithms.  Figure \ref{fig:synthetic_comparison} shows results on synthetic data that demonstrate that TVD does, inf act, imply a very different ordering of the relative performance of different learning algorithms than that implied by SHD and SID.
We began by constructing 30 random DAGs with 14 variables and $E[N]=2$.
We generated parameters on those DAGs using each of the synthetic data techniques and sampled 5,000 data points from each DAG. Then, we applied PC, MMHC, and GES to the resulting data sets and measured the SID, SHD, and sum of pairwise total variations. As shown in Figure~\ref{fig:synthetic_comparison}, some of the findings that would be reached with SID and SHD are not supported by a TV evaluation. The structural measures suggest that MMHC outperforms PC on the Dirichlet domain. However, the performance of the two algorithms is statistically indistinguishable as measured by TV. When measured with SID or SHD, GES does not outperform either MMHC or PC. However, GES is consistently the best performing algorithm in terms of interventional distribution accuracy.

Experiments in the paper demonstrate that TVD can, at least in some cases, provide information that structural measures cannot.  However, that does not mean that the additional information is useful.  To address this concern, we sought to measure how TVD responds to specific types of errors in learned structure.  Specifically, we evaluate the effects of over-specification (extraneous edges) and under-specification (omitted edges) on model performance. We used our three empirical data sets drawn from large-scale computational systems (JDK, Postgres, and HTTP) to perform this analysis. From the original exhaustive experiments, we can identify which treatment-outcome pairs are causally related.  We construct a partial DAG, consisting only of edges between treatment and outcome, by introducing an edge between each pair of causally related treatment and outcome. Then, a pseudo-observational data set can be constructed by sub-sampling treatment assignments according to a biasing covariate (details in Supplemental Materials). The resulting DAG model (illustrated for the JDK data set in Figure~\ref{fig:jdk-consistent}) consistently estimates distributions $P(O|do(T=t))$ for all treatment-outcome pairs.

We altered the consistent models of each data set to induce over-specification and under-specification. To quantify the effects of over-specification, we produced models in which one of the treatment variables had a directed edge into every outcome, regardless of the causal relationships in the true model. To quantify the effects of under-specification, we produced models in which one of the treatment variables had no outgoing edges. This process was repeated for each of our three domains and each treatment variable within that domain. For each model, a sum of pairwise total variations was computed as $\sum_{T,O} TV_{P,\hat{P},T=1}(O)$, where $P$ represents the reference distribution given by the consistent model (as in Figure~\ref{fig:jdk-consistent}) and $\hat{P}$ represents the distribution induced by the altered model. A comparison of TVD, SHD, and SID on these experiments is shown in Table~\ref{tbl:altered-models}. 

Two properties are apparent.  First, over-specification is penalized differently by different evaluation measures. For small data sets, such as the JDK domain, over-specified models have zero SID but significant TVD values due to loss of statistical efficiency.  Second, penalizing over-specification and under-specification with equal cost, as in SHD, is inconsistent with interventional distribution quality. In these domains, model under-specification has 2-5x the distributional impact of under-specification as measured by total variation.

\begin{table*}[t]
\centering
\caption{Metric Comparison on Real Domains with Over-specification and Under-specification}
{ \small
 \begin{tabular}{ lll | rrr | rrr | rrr  } 
  \hline
 Domain & Subjects & Model Type & \multicolumn{3}{c|}{SID: Min, Median, Max} & \multicolumn{3}{c|}{SHD: Min, Median, Max} & \multicolumn{3}{c}{TVD: Min, Median, Max} \\ 
  \hline
\multirow{2}{*}{JDK} & \multirow{2}{*}{473} & Over-specify & 0 & \hspace{3em} 0 & 0 & 1 & \hspace{2.5em} 3 & 3 & 0.04 & \hspace{0.75em} 0.17 & 0.21 \\ 
  &  & Under-specify & 4 & 5 & 9 & 2 & 2 & 4 & 0.22 & 0.41 & 0.58 \\ 
  \hline
  \multirow{2}{*}{Postgres} & \multirow{2}{*}{5,000} & Over-specify & 0 & 0 & 0 & 0 & 1 & 2 & 0.00 & 0.06 & 0.09 \\ 
  &  & Under-specify & 4 & 6 & 8 & 3 & 4 & 5 & 0.17 & 0.35 & 0.61 \\ 
  \hline
  \multirow{2}{*}{HTTP} & \multirow{2}{*}{2,599} & Over-specify & 0 & 0 & 0 & 1 & 2 & 4 & 0.06 & 0.06 & 0.09 \\ 
  &  & Under-specify & 2 & 6 & 10 & 1 & 3 & 4 & 0.22 & 0.25 & 0.30 \\ 
   \hline
\end{tabular}
}
\label{tbl:altered-models}
\end{table*}

\section{Additional Details on Presented Experiments}
\begin{figure}
\centering
\includegraphics[width=0.7\linewidth]{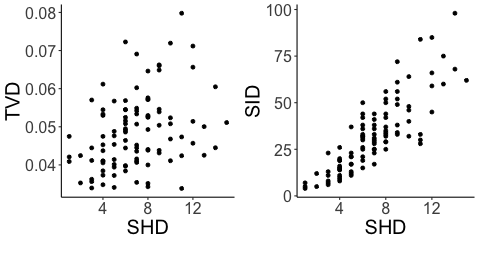}
\caption{Structural and Interventional Measures Compared on Synthetic Data with MMHC.}
\label{fig:chickering_mmhc}
\end{figure}

\begin{figure}
\centering
\includegraphics[width=0.7\linewidth]{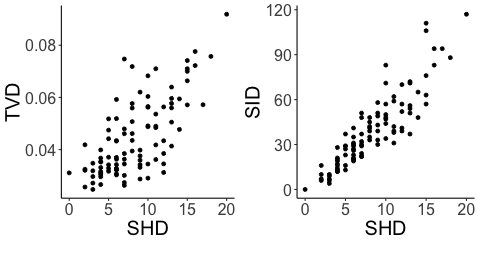}
\caption{Structural and Interventional Measures Compared on Synthetic Data with PC.}
\label{fig:chickering_pc}
\end{figure}

\begin{figure}
\centering
\includegraphics[width=0.6\linewidth]{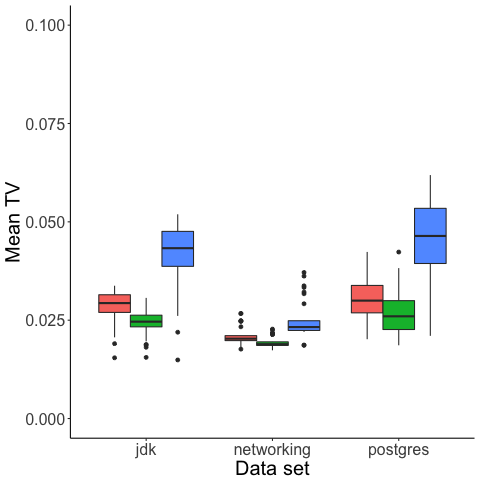}
\caption{Results for MMHC for the Experiments Described in the Paper, using Synthetic Data that has been Created to Look like Empirical Data}
\label{fig:empirical_results_mmhc}
\end{figure}

Figures \ref{fig:chickering_mmhc} and \ref{fig:chickering_pc} show the results of comparing synthetic and interventional measures on synthetic data for both MMHC and PC. (results for GES were presented in the paper) Interestingly, while the correlation between SID and SHD is relatively consistent for all three structure learning algorithms, the correlation between TVD and SHD varies substantially, from seemingly completely uncorrelated (GES) to very clearly correlated (PC).  This suggests that, in some cases, structural measures can provide a decent proxy for interventional measures.  However, it is unlikely that the researcher knows this to be the case ahead of time, and the comparative difference in TVD between the three algorithms suggests the value of using TVD when comparing multiple causal learning algorithms.
We also provide additional results for experiments discussed in the paper that created synthetic data sets by learning their structure from empirical data.  While we reported results using GES and PC, here we show results for MMHC.  Figure \ref{fig:empirical_results_mmhc} shows the performance of three learning algorithms (GES, MMHC, and PC).  MMHC was used to infer a causal model from empirical data, and that model was then used to generate the synthetic data.  Compared with the results in the paper, the relative performance of different algorithms looks somewhat similar to the results using GES, though there are some differences (e.g., PC is clearly the worst on all data sets in Figure \ref{fig:empirical_results_mmhc}, while this is not the case for GES in Figure 1 in the paper).

Sample sizes for some of the software system data sets are small, so in Figure \ref{fig:empirical_results_mmhc} and Figure 1 in the paper, we report results as distributions over 30 trials for each algorithm and data set.  

\nocite{huang2018generalized} \nocite{wu2018discrimination} \nocite{agrawal2018minimal} \nocite{ghassami2017budgeted} \nocite{janzing2018detecting} \nocite{yabe2018causal} \nocite{yang2018characterizing} \nocite{cai2018self} \nocite{yeo2018machine} \nocite{hao2018measuring} \nocite{zhang2018fairness} \nocite{alrajeh2018combining} \nocite{rubenstein2017probabilistic} \nocite{zhang2017causal1} \nocite{zhalama2017sat} \nocite{soleimani2017treatment} \nocite{gong2017causal} \nocite{kilbertus2017avoiding} \nocite{ambrogioni2017gp} \nocite{ghassami2017learning} \nocite{wang2017permutation} \nocite{louizos2017causal} \nocite{kocaoglu2017experimental} \nocite{kocaoglu2017entropic} \nocite{zhang2017causal2} \nocite{kruengkrai2017improving} \nocite{correa2017causal} \nocite{achab2017uncovering} \nocite{chaudhry2017uncertainty} \nocite{kansky2017schema} \nocite{chalupka2016unsupervised} \nocite{pena2016alternative} \nocite{lee2016characterization} \nocite{arbour2016inferring} \nocite{zhang2016identifiability} \nocite{kallus2016causal} \nocite{asbeh2016pairwise} \nocite{roumpelaki2016marginal} \nocite{triantafillou2016score} \nocite{lattimore2016causal} \nocite{toulis2016long} \nocite{zhou2016causal} \nocite{magliacane2016ancestral} \nocite{cheng2016ranking} \nocite{borboudakis2016towards} \nocite{kummerfeld2016causal} \nocite{merck2016causal} \nocite{lee2016learning} \nocite{xu2016learning} \nocite{affeldt2015robust} \nocite{chalupka2015visual} \nocite{marazopoulou2015learning} \nocite{mooij2015empirical} \nocite{didelez2015causal} \nocite{bareinboim2015bandits} \nocite{rothenhausler2015backshift} \nocite{gao2015local} \nocite{shanmugam2015learning} \nocite{plis2015rate} \nocite{hill2015measuring} \nocite{stanton2015mining} \nocite{chikhaoui2015new} \nocite{sun2015causal} \nocite{xu2015large} \nocite{hashimoto2015generating} \nocite{zhang2015multi} \nocite{bareinboim2015recovering} \nocite{shajarisales2015telling} \nocite{lopez2015towards} \nocite{gong2015discovering} \nocite{geiger2015causal} \nocite{scholkopf2015removing} \nocite{geiger2014estimating} \nocite{hyttinen2014constraint} \nocite{arbour2014propensity} \nocite{meek2014toward} \nocite{silva2014causal} \nocite{irfan2014causal} \nocite{hu2014randomized} \nocite{kpotufe2014consistency} \nocite{bareinboim2014recovering} \nocite{LeeB18} \nocite{CaiQZZH18} \nocite{FigueiredoBMA18} \nocite{MagliacaneOCBVM18} \nocite{ZhangB18a} \nocite{WangSBU18} \nocite{LeeCO18} \nocite{HuCNCG18} \nocite{LindgrenKDV18} \nocite{GhassamiKHZ18} \nocite{KallusMU18} \nocite{MitrovicST18} \nocite{KurutachTYRA18} \nocite{ShermanS18} \nocite{AcharyaBDK18} \nocite{BelloH18a}
 \nocite{MogensenMH18} \nocite{ShpitserS18} \nocite{BlomKMM18} \nocite{ZhangB18b} \nocite{NabiKS18} \nocite{SubbaswamyS18} \nocite{JaberZB18} \nocite{0001GRBSG18} \nocite{RubensteinBMS18} \nocite{CuiGSH18} \nocite{ForreM18} \nocite{JavidianV18} \nocite{WolfeSF18} \nocite{Pena18} \nocite{HensleyMMBHL18} \nocite{Wang18} \nocite{NabiMS18} \nocite{Pouliot18} \nocite{NabiS18} \nocite{BlomM18} \nocite{BenF18} \nocite{BongersM18} \nocite{NieW18} \nocite{NarayanLE18} \nocite{ArmenE18} \nocite{HowardRMS18}

\renewcommand{\refname}{\normalsize References}
\bibliography{causal-eval.bib}
\bibliographystyle{plainnat}